%% file: icorr17.tex
\documentclass[letterpaper, 10 pt, conference]{ieeeconf}  

\IEEEoverridecommandlockouts                              

\makeatletter
\let\NAT@parse\undefined
\makeatother
\usepackage[numbers,sort,comma,square]{natbib}
\usepackage{amsmath}
\usepackage{tikz}                      
\usepackage{pgfplots}                  
\usepackage[caption=false]{subfig}     
\usepackage[pdftex,colorlinks=false,bookmarksopen=true,bookmarksopenlevel=0,pdfborder={0 0 0},breaklinks]{hyperref}
\usepackage[nolist]{acronym}           
\usepackage{url}
\usepackage{bm}
\usepackage{booktabs}                  
\usepackage{flexisym}
\usepackage{amsfonts}
\usepackage{soul}
\usepackage[switch]{lineno}

\input{Configuration}

\input{Macro}

\title{\LARGE \bf
Adaptive Learning to Speed-Up Control of Prosthetic Hands: a Few Things Everybody Should Know%
}

\author{Valentina~Gregori$^{1}$, Arjan~Gijsberts$^{1}$ and Barbara~Caputo$^{1}$
\thanks{*This work was supported by the Swiss National Science Foundation Sinergia project \#160837 ``Megane Pro''}
\thanks{$^{1}$Department of Computer, Control, and Management Engineering, University of Rome La Sapienza, via Ariosto 25, 00185 Roma, Italy 
        {\tt\small surname@dis.uniroma1.it}}%
}

\begin{document}

\maketitle
\thispagestyle{empty}
\pagestyle{empty}

\input{Abstract}
\input{Introduction}

\input{RelatedWork}
\input{Algorithms}
\input{ExperimentalSetup}
\input{Experiments}
\input{Conclusions}



{\small
\bibliographystyle{IEEEtranN} 
\bibliography{icorr17}%
}

\end{document}

%% file: Configuration.tex
\begin{acronym}
  \acro{CV}{cross validation}
  \acro{DWT}{Discrete Wavelet Transform}
  \acro{HIST}{\ac{SEMG} Histogram}
  \acro{LSSVM}[LS-SVM]{Least-Squares Support Vector Machine}
  \acro{MAV}{Mean Absolute Value}
  \acro{VAR}{Variance}
  \acro{WL}{Waveform Length}
  \acro{MDWT}[mDWT]{marginal \acl{DWT}}
  \acro{MEAN}{mean value}
  \acro{MKAL}{Multi Kernel Adaptive Learning}
  \acro{MULTIKT}[MultiKT]{Multi Model Knowledge Transfer}
  \acro{NINAPRO}[NinaPro]{Non-Invasive Adaptive Prosthetics}
  \acro{NMSE}[nMSE]{normalized Mean Squared Error}
  \acro{NOTRANSFER}[NoTransfer]{no transfer model}
  \acro{PRIOR}[Prior]{prior model}
  \acro{PRIORLIN}[Prior]{prior model}
  \acro{PRIORRBF}[PriorRBF]{prior RBF model}
  \acro{PSD}{power spectral density}
  \acro{RBF}{Radial Basis Function}
  \acro{RMS}{Root Mean Square}
  \acro{SEMG}[sEMG]{surface electromyography}
  \acro{SVM}{Support Vector Machine}
  \acro{HTL}{Hypothesis Transfer Learning}
\end{acronym}

\usetikzlibrary{plotmarks}
\usetikzlibrary{external}

\pgfkeys{
  /pgf/number format/set thousands separator={},    
}

\pgfplotsset{
  compat=newest,
  every axis/.append style={
    font=\footnotesize,
    line width=0.5pt,
  },
  ylabel near ticks,  
  xlabel near ticks,
  legend style={
    font=\footnotesize,
    draw=none,
    /tikz/every odd column/.append style={column sep=2pt},
    /tikz/every even column/.append style={column sep=0.5cm},
  },
  legend columns=-1,
  legend style={at={(0.5,1.02)},anchor=south},
  ymajorgrids,
  grid style={color=black!10},
  error bars/y dir=both,
  multikt/.style={cyan,solid,mark=triangle*,mark options={solid},mark size=1.6pt,},
  priorlin/.style={black,solid,mark=diamond*,mark options={solid},mark size=1.6pt,},
  priorrbf/.style={red!40!white,solid,mark=heart,mark options={solid},mark size=1.6pt,},
  notransfer/.style={red,solid,mark=*,mark options={solid},mark size=1.6pt,},
  mkal/.style={black!60!green,solid,mark=x,mark options={solid},mark size=1.6pt,},
}



%% file: Abstract.tex
\begin{abstract}
A number of studies have proposed to use domain adaptation to reduce the training efforts needed to control an upper-limb prosthesis exploiting pre-trained models from prior subjects. 
These studies generally reported impressive reductions in the required number of training samples to achieve a certain level of accuracy for intact subjects.
We further investigate two popular methods in this field to verify whether this result equally applies to amputees. 
Our findings show instead that this improvement can largely be attributed to a suboptimal hyperparameter configuration.
When hyperparameters are appropriately tuned, the standard approach that does not exploit prior information performs on par with the more complicated transfer learning algorithms. 
Additionally, earlier studies erroneously assumed that the number of training samples relates proportionally to the efforts required from the subject. 
However, a repetition of a movement is the atomic unit for subjects and the total number of repetitions should therefore be used as reliable measure for training efforts. 
Also when correcting for this mistake, we do not find any performance increase due to the use of prior models.
\end{abstract}

%% file: Introduction.tex
\section{Introduction}\label{sec:introduction}

A majority of upper-limb amputees is interested in prostheses controlled via \ac{SEMG}, but they perceive the difficult control as a great concern~\cite{atkins96}. 
Machine learning has opened a new path to tackle this problem by allowing the prosthesis to adapt to the myoelectric signals of a specific user. 
Although these methods have been applied with success in an academic setting (e.g., \cite{castellini09} and references therein), they require a long and painful training procedure to learn models with satisfactory performance.

Several studies have proposed to reduce the amount of required training data by leveraging over previous models from different subjects~\cite{orabona09, tommasi11,patricia14}. 
The underlying idea is that a model for a new target user can be bootstrapped from a set of prior source models. 
Though the idea is appealing and initial studies have shown remarkable improvements, there is no conclusive evidence that these strategies lead to tangible benefits in the real world.
An obvious limitation in the earlier studies is that they only considered intact subjects. 
This is relevant since myoelectric signals are user-dependent and this holds in particular for amputees, as the amputation and subsequent muscular use have a considerable impact on the quality on the myoelectric signals~\cite{farina02}.

Other limitations relate to the technical and conceptual execution of the experimental validation.
First, the hyperparameters of the algorithms were not optimized for the method at hand, but rather chosen based on how well they performed on average when applied on the data of other subjects.
Individual hyperparameter optimization for the methods and number of training samples is crucial for the successful application of machine learning algorithms and omission of this procedure may skew results.
For instance, this tuning procedure may give an unfair disadvantage to the baseline that does not use prior information from pre-trained users.

On the conceptual level, the previous studies used the number of training samples (i.e., windows of the myoelectric signals) as measure for the required training effort.
A consequence of this strategy is that not all available training samples were used to classify the movements, since subjects cannot produce individual samples. 
Instead, a repetition of a movement consisting of multiple windows is the atomic unit for subjects.
This artificial reduction of training information is likely to be disadvantageous for the baseline that only relies on target training data.
In our evaluation, we instead use the number of movement repetitions as realistic measure of the training effort.
Furthermore, we consider all possible data, subjects and combinations of training repetitions, thereby removing the possible effects of random selection from the evaluation.

In this paper, we provide more insight into the benefits of domain adaptation for prosthetic control by augmenting the experiments by \citet{patricia14} with amputated subjects while also addressing other limitations. 
This results in three experimental settings, namely (1) the original experiments according to the setup common in literature~\cite{orabona09,tommasi11,patricia14} extended with amputated subjects, (2) the same setup with hyperparameter optimization and finally (3) a realistic setup where we also address the conceptual issues. 
In each setting, we perform three experiments, where intact and amputated subjects make up the groups of target and source subjects.

This paper is structured as follows.
In \autoref{sec:relatedwork} we present the related work on domain adaptation in the context of myoelectric prosthetics. 
The algorithms that will be considered in our experiments will then be explained in detail in \autoref{sec:algorithms}. 
We continue with our experimental setup in \autoref{sec:setup}, after which we will present the results in \autoref{sec:experiments}.
Finally, we conclude the paper in \autoref{sec:conclusions}.

%% file: RelatedWork.tex
\section{Related Work}\label{sec:relatedwork}

One of the first attempts to classify myoelectric signals of three volunteers was by \citet{graupe75}. 
In the following years, studies on prosthetic control led to many advances in the analysis and understanding of \ac{SEMG}. 
\citet{castellini09} noted that myoelectric signals differ significantly from person to person and that models trained for different subjects are therefore not automatically reusable. However, they showed that a pre-trained model could be used to classify samples from similar subjects.

Several studies continued in this direction with different strategies to build more robust models that take advantage of past information from source subjects or, in the context of repeatability, from the target itself.
\citet{matsubara11} proposed to separate myoelectric data in user-dependent and motion-dependent components, and to reuse models by quickly learning just the user-dependent component for new subjects.
\citet{sensingero09} presented different methods based on an appropriate concatenation of target and source data. 
Others still approached the problem by searching for a mapping to project data from different subjects into a common domain~\cite{chattopadhyay11,khushaba14}; a similar strategy was also used to reduce the recalibration time for a target that attempts to use the prosthesis on different days~\cite{liu16towards}.

Other studies proposed to leverage over prior models from already trained source subjects without requiring direct access to their data~\cite{orabona09,tommasi11,patricia14,liu16}. They tested different types of so-called \ac{HTL} algorithms showing a gain in performance with respect to non-adaptive baselines. 
We are particularly interested in the findings of \citet{tommasi11,patricia14}, who worked with a significant number of classes and intact subjects from the public \ac{NINAPRO} database~\cite{atzori14}.
They report that the number of training samples required to obtain a given level of performance can be reduced by an order of magnitude as compared to learning from scratch.


%% file: Algorithms.tex
\section{Algorithms} \label{sec:algorithms}

We first describe the mathematical background by means of a base learning algorithm in \autoref{sec:algorithms:background}, then we proceed with the domain adaptation methods included in our evaluations in \autoref{sec:algorithms:adaptive}.

\subsection{Background}\label{sec:algorithms:background}
Let us define a training dataset $D = \left\{ \bm{x}_{i}, y_i \right\}_{i=1}^N$ of $N$ input samples $\bm{x}_i \in \mathcal{X} \subseteq \mathbb{R}^d$ and corresponding labels $y_i \in \mathcal{Y} = \left\{1, \ldots, G\right\}$. In the context of myoelectric classification, the inputs are the myoelectric signals and the labels are the movements chosen from a set of $G$ possible classes.
The goal of a classification algorithm is to find a function $h(\bm{x})$ that, for any future input vector $\bm{x}$, can determine the corresponding output $y$.
Among the algorithms that construct such a model, \acp{SVM} are some of the most popular.

The base of the domain adaptation algorithms described later on is the \ac{LSSVM}~\cite{suykens02}, a variant of \ac{SVM} with a squared loss and equality constraint. It writes the output hypothesis as $h(\bm{x}) = \langle\bm{w}, \phi(\bm{x}) \rangle + b $, where $\bm{w}$ and $b$ are the parameters of the separating hyperplane between positive and negative samples. The optimal solution is thus given by
\begin {equation}
\label{eq:ls-svm}
\begin{aligned}
&\underset{\bm{w},b}{\text{min}} \left\{ \dfrac{1}{2} \Vert \bm{w} \Vert^{2} + \dfrac{C}{2} \sum\limits_{i=1}^N \xi_{i}^{2} \right\} \\
&\text{s.t.}\quad  y_{i} = \langle\bm{w},\phi(\bm{x}_{i}) \rangle+b+\xi_{i}, \quad \forall \textit{i} \in \{ 1,...,N \}\enspace,\\
\end{aligned}\end {equation}
where $C$ is a regularization parameter and $\bm{\xi}$ denotes the prediction errors. 
We approach our multi-class classification problem via a one-vs-all scheme to discriminate each class from all others.
To obtain a better solution we mapped the input vectors $\bm{x}_i$ into a higher dimensional feature space using $\phi(\bm{x}_i)$.  
Usually, this mapping $\phi(\cdot)$ is unknown and we work directly with the kernel function $K(\bm{x}\textprime, \bm{x})~=~\langle \phi(\bm{x}\textprime), \phi(\bm{x}) \rangle$~\cite{suykens02}.
In the following, we use a \ac{RBF} kernel
\begin {equation}
\label{eq:GausKer}
K(\bm{x}\textprime, \bm{x}) = e^{- \gamma \parallel \bm{x}\textprime - \bm{x} \parallel^{2}}\enspace \text{with}\enspace \gamma > 0\enspace.
\end {equation}

\subsection{Adaptive Learning}\label{sec:algorithms:adaptive}
Domain adaptation algorithms construct a classification model for a new target using past experience from the sources.
More specifically, let us assume that we have $K$ different sources, where each source is a classification model for the same set of movements.
The used \ac{HTL} algorithms can then be described as follows.

\subsubsection{\acl{MULTIKT}}\label{sec:algorithms:multikt}
This method aims to find a new separating hyperplane $\bm{w}$ that is close to a linear combination of the pre-trained source hypotheses $\hat{\bm{w}}^{k}$~\cite{tommasi11,tommasi14}. We solve the optimization problem
\begin{equation}
\label{eq:OptimMultiAdaMulti}
\begin{aligned}
&\underset{\bm{w},b}{\text{min}}\> \left\{ \dfrac{1}{2} \Big\Vert \bm{w}-\sum\limits_{k=1}^K \beta^{k} \hat{\bm{w}}^{k} \Big\Vert^{2}+\dfrac{C}{2} \sum\limits_{i=1}^N \xi_{i}^{2} \right\} \\
&\text{s.t.}\quad y_{i} = \langle\bm{w}, \phi(\bm{x}_{i})\rangle + b + \xi_{i}\enspace.
\end{aligned}
\end{equation}
The vector $\bm{\beta} = \left[\beta_{1},...,\beta_{K}\right]^{T}$ with $\beta_k \geq 0$ and $\Big\Vert \bm{\beta} \Big\Vert^{2}~\leq~1$ represents the contribution of each source in the target problem and is obtained by optimizing a convex upper bound of the leave-one-out misclassification loss~\cite{tommasi14}.
A more general case consists of different weights for different classes of the same source, such that $\beta_{k,g}$ is the weight associated to class $g$ of source $k$. 
In this work we used this latter version of the algorithm.

\subsubsection{\acl{MKAL}}\label{sec:algorithms:mkal}
This algorithm combines source and target information via a linear combination of kernels~\cite{orabona10, orabona12}.  
Let us define
\begin{align}
\label{eq:WBar_PhiBar}
\bar{\bm{w}} & = [\bm{w}^{0},\bm{w}^{1},...,\bm{w}^{K}] \quad \text{and} \\
\bar{\phi}(\bm{x},y) & = [\phi^{0}(\bm{x},y),\phi^{1}(\bm{x},y),...,\phi^{K}(\bm{x},y)]\enspace,
\end{align}
respectively as the concatenation of the target and source hyperplanes and the mapping functions into the corresponding feature spaces.
These are both composed of $(K + 1)$ elements: the first refers to the target and the remaining ones to the sources. 
The optimization problem becomes
\begin{equation}
\label{eq:MinProb_MKAL}
\begin{aligned}
&\underset{\bar{\bm{w}}}{\text{min}} \left\{ \dfrac{\lambda}{2} \parallel \bar{\bm{w}} \parallel^{2}_{2,p} + \dfrac{1}{N} \sum\limits_{i=1}^N \xi_{i} \right\}\\
&\text{s.t.}\quad \langle \bar{\bm{w}} , (\bar{\phi} (\bm{x}_{i},y_{i}) - \bar{\phi} (\bm{x}_{i},y) ) \rangle \geq 1 - \xi_{i},\, \forall \textit{i} \  y \neq y_{i}\enspace.
\end{aligned}
\end{equation}
The element $p$ regulates the level of sparsity in the solution $\bar{\bm{w}}$ and can vary in the range $(1, 2]$. The solution is obtained via stochastic gradient descent during $T$ epochs over the shuffled training samples.

%% file: ExperimentalSetup.tex
\section{Experimental Setup}\label{sec:setup}

The experimental evaluation is subdivided in three settings. 
The first one is modeled after related literature for this kind of experiments with \ac{SEMG} data~\cite{orabona09,tommasi11,patricia14}. 
The second is identical but adds hyperparameter optimization for the target models.
The third and final one is a novel framework in which we fixed the shortcomings of the previous settings to make the experiments as realistic as possible.
We will refer to the settings as original, optimized and realistic.
In the following, we first explain the used data and classifiers, and subsequently elaborate on the experimental settings.

\subsection{Data}\label{sec:setup:data}
The data used in our work are from the \ac{NINAPRO} database\footnote{\url{http://ninapro.hevs.ch/}}~\cite{atzori14}, the largest publicly available database for prosthetic movement classification with 40 intact subjects and 11 amputees. 
Each subject executed 40 movements for 6 times, such that each repetition was alternated with a rest posture.
While performing the movements, twelve electrodes acquired \ac{SEMG} data from the arm of the subject.
The standardized data were used according to the control scheme by \citet{englehart03}, where we extracted features from a sliding window of 200\,ms and an increment of 10\,ms.
The resulting set of windows was subsequently split in train and test sets for the classifier; data from repetitions $\left\{ 1, 3, 4, 6 \right\}$ were dedicated to training while data from repetitions 2 and 5 were used as test.
To reduce the computational requirements, we subsampled the training data by a factor of 10 at regular intervals.

\subsection{Classifiers}\label{sec:setup:classifiers}

The algorithms used to build the classification models were the two mentioned \ac{HTL} algorithms together with two baselines:
\begin{itemize}
\item the \emph{\ac{NOTRANSFER}}, which uses an \ac{LSSVM} with \ac{RBF} kernel trained only on the target data. This corresponds to learning without the help of prior knowledge.
\item the \emph{\ac{PRIOR}}, which learns an \ac{LSSVM} with linear kernel on top of the raw predictions of the source models. This measures the relevance of the source hypotheses by using them as feature extractors for the target data.
\item \emph{\ac{MULTIKT}}, as explained in \autoref{sec:algorithms:multikt}, which learns a model on the target data that is close to a weighted combination of the source hypotheses.
\item \emph{\ac{MKAL}}, as explained in \autoref{sec:algorithms:mkal}, which linearly combines an \ac{RBF} kernel on the target data with the source predictions. Parameters \textit{p} and \textit{T} were set to 1.04 and 300.
\end{itemize}
The classification models for the sources were based on a non-linear \ac{SVM} or \ac{LSSVM} with \ac{RBF} kernel. 

\subsection{Settings}\label{sec:setup:settings}

For each of the settings, we ran three experiments with distinct groups of target and source subjects:
\begin{itemize}
\item Intact-Intact: intact target subjects exploit prior knowledge of other intact sources;
\item Amputees-Amputees: amputated target subjects exploit previous experience of other amputees;
\item Amputees-Intact: amputated target subjects exploit prior knowledge of intact subjects.
\end{itemize}
In the first and second experiment, each subject toke the role of target just once, while the remaining subjects were used as sources.
In the third, all of the amputees were once the target and the set of intact subjects was used only as sources.

\subsubsection{Original Setting}\label{sec:setup:settings:original}

The purpose of the original and optimized settings is to investigate the isolated impact of hyperparameter optimization on the performance of the target classifiers.
We therefore replicated, as closely as possible, the experiments from \citet{patricia14} with 9 amputees\footnote{We omitted two amputees from the database that had only 10 electrodes due to insufficient space on their stump.}. and a random subset of 20 intact subjects from the \ac{NINAPRO} database. For these subjects we considered 17 movements plus the rest posture, appropriately subsampled to balance it with the other movements.    
The \ac{SEMG} representation used in this setting was the average of \ac{MAV}, \ac{VAR} and \ac{WL} features~\cite{kuzborskij12}, as to reduce the dependency on one specific type of representation.
The details of this and the subsequent settings are presented schematically in \autoref{tab:setup:settings}.
\begin{table*}
  \caption{Experimental settings.}
  \centering%
  \begin{tabular} {l p{2.0cm} c l l p{3.0cm} p{3.0cm}}
   	\textbf{Setting}   & \textbf{Movements} & \# \textbf{Int./Amp.} & \textbf{Features} & \textbf{Source model} & \textbf{Source hyperparameters} & \textbf{Target hyperparameters} \\
   	\midrule
    \textbf{Original}  & 9 wrist, 8 finger, subsampled rest & 20/9 & avg. \acs{MAV}/\acs{VAR}/\acs{WL} & \acs{SVM} & balanced accuracy on other subjects & same as source \\
  	\addlinespace
    \textbf{Optimized} & 9 wrist, 8 finger, subsampled rest & 20/9 & avg. \acs{MAV}/\acs{VAR}/\acs{WL} & \acs{SVM} & balanced accuracy on other subjects & balanced accuracy of 5-fold \acs{CV} over shuffled training samples of target\\
  	\addlinespace
    \textbf{Realistic} & 9 wrist, 8 finger, 23 grasp & 40/8 & \acs{MDWT} & \acs{LSSVM} & accuracy of k-fold \acs{CV} over repetitions of source & accuracy of k-fold \acs{CV} over train repetitions of target \\
	\end{tabular}
	\label{tab:setup:settings}
\end{table*}

The source models were created by training an \ac{SVM} with \ac{RBF} kernel using all training repetitions of the respective subject.
For the target models we trained the classifiers on an increasing number of random samples, from 120 to 2160 in steps of 120, from the training repetitions. 
The hyperparameters for both the source and target models were chosen from $C, \gamma \in \lbrace 0.01,0.1,1,10,100,1000 \rbrace$ and kept constant regardless of the number of training samples.
For each parameter configuration, we evaluated the average balanced classification accuracy of each source subject when tested on the target subjects. 
For the target subject and its source models, we then chose the configuration that maximizes this average, making sure to exclude the data from the target subject.
The motivation for this procedure is that biased regularization in \ac{MULTIKT} requires the source and target models to ``live'' in the same space.

\subsubsection{Optimized Setting}\label{sec:setup:settings:hyper}

Strictly speaking, the above assumption only requires that all the sources have the same \ac{RBF} bandwidth $\gamma$ as the related target.
Moreover, \ac{MULTIKT} can also be interpreted as predicting the difference between the source predictions and the true labels~\cite{kuzborskij16}. 
In this alternative interpretation, there is no need for source and target models to use the same kernel. 
We therefore tuned the hyperparameters in the optimized setting for each individual target and for each training set size based on 5-fold \ac{CV} on the target training set.
Note, however, that we still use the original method to determine the parameters for the source models.

\subsubsection{Realistic Setting}\label{sec:setup:settings:realistic}

In the final setting we extended the hyperparameter optimization procedure and attempted to address all issues to make the experiments as realistic as possible. 
First, we considered all available movements and subjects in the \ac{NINAPRO} database\footnote{Data for the first amputated subject was omitted, since the acquisition was interrupted prematurely.}.
As \ac{SEMG} representation we used \ac{MDWT} features, which have previously shown excellent performance in related work on this database~\cite{gijsberts14}.

The main conceptual innovation with respect to the previous settings is that we trained target models on an increasing number of repetitions. 
The motivation is that the effort of the subject during data acquisition is given by the required number of repetitions of each movements, so we analyze the accuracy as a function of this atomic unit.
Given the set of training repetitions $\left\{ 1, 3, 4, 6 \right\}$, we considered all possible subsets of length between 1 and 4 repetitions.
For all these cases, we optimized the target model using $k$-fold \ac{CV}, where each fold corresponded to samples belonging to one repetition.
In the exceptional case of only a single training repetition, we instead used 5-fold \ac{CV} over the samples.
The parameter grid was extended to $C \in \lbrace 2^{-6}, 2^{-4}, \ldots, 2^{12}, 2^{14} \rbrace$ and $\gamma \in \lbrace 2^{-20}, 2^{-18}, \ldots, 2^{-2}, 2^{0} \rbrace$.
Models for the source subjects on the other hand were trained using all repetitions and the hyperparameters were optimized specifically for the individual subject using 6-fold \ac{CV}, where the folds again corresponded to the repetitions.
The source models were built with \ac{LSSVM} instead of \ac{SVM} to be more coherent with the other classifiers, which are all derived from \ac{LSSVM}.
Due to the much larger number of samples in this realistic setting\footnote{Each repetition consists of approximately 35000 samples.} we further subsampled the data used for hyperparameter optimization by a factor of 4. For the same reason we decided to omit \ac{MKAL} from the analysis.

%% file: Experiments.tex
\section{Experiments} \label{sec:experiments}

In this section we first investigate the isolated impact of hyperparameter optimization when applied to the original setting. 
Then we verify whether the findings also apply to the realistic setting described in \autoref{sec:setup:settings:realistic}. 
An in-depth discussion follows on the explanations of the results.

\subsection{Results}\label{sec:experiments:hyper}

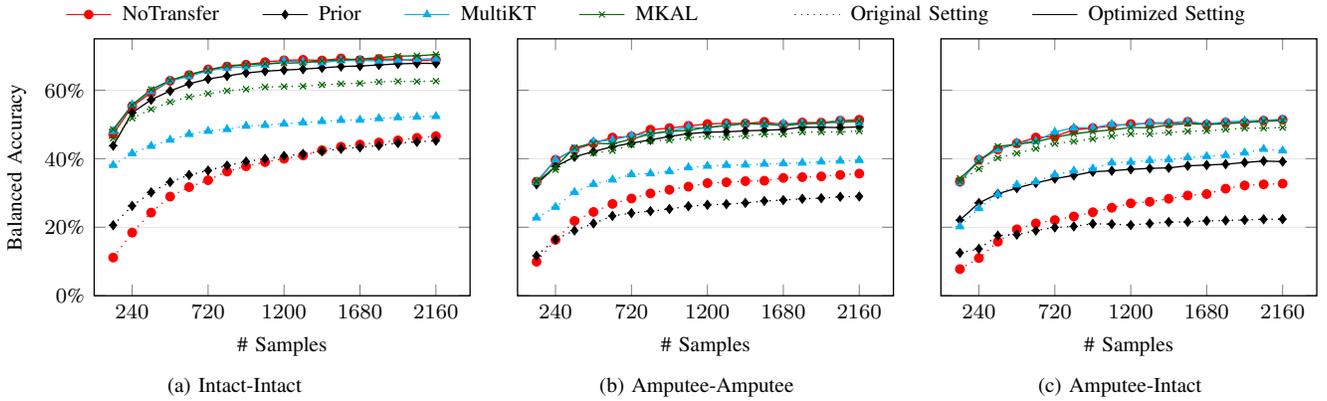
\begin{figure*}[tbp]
  \pgfplotsset{%
    yticklabel={\pgfmathparse{\tick*100}\pgfmathprintnumber{\pgfmathresult}\%},
    xtick={240,720,1200,1680,2160},
    height=5cm,
  	width=0.37\linewidth,
    xlabel=\# Samples,
    ymin=0, ymax=0.75,
    xmin=0,
    original/.style={dotted},
    hyper/.style={solid},
  }
  \pgfplotstableset{
    every table/.append style={x=samples,y=mean,y error=std},
  }
  \centering%
  \ref{fig:experiments:hyper:legend}\\[-3mm]
  \subfloat[Intact-Intact] {%
    \label{fig:experiments:hyper:intactintact}%
    \tikzsetfigurename{adapt_optimized_II}
    \begin{tikzpicture}
      \begin{axis}[ylabel=Balanced Accuracy]
        \addplot+[notransfer,hyper] table[] {acc_no_II_hyper.txt};
        \addplot+[priorlin,hyper] table[] {acc_prior_II_hyper.txt};
        \addplot+[multikt,hyper] table[] {acc_multikt_II_hyper.txt};
        \addplot+[mkal,hyper] table[] {acc_tl_II_hyper.txt};

        \addplot+[notransfer,original] table[] {acc_no_II_original.txt};
        \addplot+[priorlin,original] table[] {acc_prior_II_original.txt};
        \addplot+[multikt,original] table[] {acc_multikt_II_original.txt};
        \addplot+[mkal,original] table[] {acc_tl_II_original.txt};
      \end{axis}
    \end{tikzpicture}
  }%
  \hfill%
  \subfloat[Amputee-Amputee] {%
    \label{fig:experiments:hyper:amputeeamputee}%
    \tikzsetfigurename{adapt_optimized_AA}
    \begin{tikzpicture}
      \begin{axis}[yticklabel=\empty]
        \addplot+[notransfer,hyper] table[] {acc_no_AA_hyper.txt};
        \addplot+[priorlin,hyper] table[] {acc_prior_AA_hyper.txt};
        \addplot+[multikt,hyper] table[] {acc_multikt_AA_hyper.txt};
        \addplot+[mkal,hyper] table[] {acc_tl_AA_hyper.txt};

        \addplot+[notransfer,original] table[] {acc_no_AA_original.txt};
        \addplot+[priorlin,original] table[] {acc_prior_AA_original.txt};
        \addplot+[multikt,original] table[] {acc_multikt_AA_original.txt};
        \addplot+[mkal,original] table[] {acc_tl_AA_original.txt};
      \end{axis}
    \end{tikzpicture}
  }%
  \hfill%
  \subfloat[Amputee-Intact] {%
    \label{fig:experiments:hyper:amputeeintact}%
    \tikzsetfigurename{adapt_optimized_AI}
    \begin{tikzpicture}
      \begin{axis}[yticklabel=\empty,legend to name=fig:experiments:hyper:legend]
        \addplot+[notransfer,hyper] table[] {acc_no_AI_hyper.txt};
        \addlegendentry{\acs{NOTRANSFER}}
        \addplot+[priorlin,hyper] table[] {acc_prior_AI_hyper.txt};
        \addlegendentry{\acs{PRIORLIN}}
        \addplot+[multikt,hyper] table[] {acc_multikt_AI_hyper.txt};
        \addlegendentry{\acs{MULTIKT}}
        \addplot+[mkal,hyper] table[] {acc_tl_AI_hyper.txt};
        \addlegendentry{\acs{MKAL}}

        \addlegendimage{empty legend}%
        \addlegendentry{}%
        \addlegendimage{black,original}%
        \addlegendentry{Original Setting}%
        \addlegendimage{black,hyper}%
        \addlegendentry{Optimized Setting}%

        \addplot+[notransfer,original] table[] {acc_no_AI_original.txt};
        \addplot+[priorlin,original] table[] {acc_prior_AI_original.txt};
        \addplot+[multikt,original] table[] {acc_multikt_AI_original.txt};
        \addplot+[mkal,original] table[] {acc_tl_AI_original.txt};
      \end{axis}
    \end{tikzpicture}
  }%
  \caption{Balanced classification accuracy for \ac{MULTIKT}, \ac{MKAL}, \ac{NOTRANSFER} and \ac{PRIORLIN} in the original and hyperparameter optimized settings.}%
  \label{fig:experiments:original}%
\end{figure*}

In \autoref{fig:experiments:original} we report the balanced classification accuracy as a function of the number of training samples averaged over all target subjects.   
The dotted lines indicate the results obtained in the original experimental framework usually employed in literature (see \autoref{sec:setup:settings:original}). 
As in the related studies, \ac{MKAL} and \ac{MULTIKT} outperform the baselines \ac{NOTRANSFER} and \ac{PRIOR} by a significant margin for all training set sizes.
This has led to the claim that the adaptive algorithms can achieve similar performance as \ac{NOTRANSFER} using an order of magnitude less training samples.
Since this improvement is observed whether the target and source subjects are intact or amputated, it is also assumed that amputees can equally exploit prior information from intact as well as other amputated subjects.

When looking at the solid lines in \autoref{fig:experiments:original}, which show results with hyperparameter optimization, we observe that the discrepancies between the algorithms disappear. 
In other words, the \ac{NOTRANSFER} baseline performs just as well as or even slightly better than the adaptive algorithms.
Furthermore, with hyperparameter optimization all methods now outperform the results in the original setting.
The only exception to this observation is \acs{PRIORLIN}, which has lower accuracy in the Amputee-Intact experiment. 
Contrary to the earlier statements, this demonstrates that prior models from intact subjects are not as useful as those from other amputees.
Together with the observation that \ac{MKAL} and \ac{MULTIKT} perform nearly identically to \ac{NOTRANSFER}, this also allows us to conclude that rather than transferring from prior models, the \ac{HTL} algorithms rely almost exclusively on target data.

\begin{figure*}[tbp]
  \pgfplotsset{%
    yticklabel={\pgfmathparse{\tick*100}\pgfmathprintnumber{\pgfmathresult}\%},
    xtick={1,2,3,4},
    height=5cm,
  	width=0.37\linewidth,
    xlabel=\# Repetitions,
    ymin=0.45, ymax=0.75,
  }
  \pgfplotstableset{
    every table/.append style={x=repetitions,y=mean,y error=std},
  }
  \centering%
  \ref{fig:experiments:realistic:legend}\\[-3mm]
  \subfloat[Intact-Intact] {%
    \label{fig:experiments:realistic:intactintact}%
    \tikzsetfigurename{adapt_realistic_II}
    \begin{tikzpicture}
      \begin{axis}[ylabel=Accuracy]
        \addplot+[notransfer] table[] {acc_no_II_realistic.txt};
        \addplot+[priorlin] table[] {acc_prior_II_realistic.txt};
        \addplot+[multikt] table[] {acc_multikt_II_realistic.txt};
      \end{axis}
    \end{tikzpicture}
  }%
  \hfill%
  \subfloat[Amputee-Amputee] {%
    \label{fig:experiments:realistic:amputeeamputee}%
    \tikzsetfigurename{adapt_realistic_AA}
    \begin{tikzpicture}
      \begin{axis}[yticklabel=\empty,]
        \addplot+[notransfer] table[] {acc_no_AA_realistic.txt};
        \addplot+[priorlin] table[] {acc_prior_AA_realistic.txt};
        \addplot+[multikt] table[] {acc_multikt_AA_realistic.txt};
      \end{axis}
    \end{tikzpicture}
  }%
  \hfill%
  \subfloat[Amputee-Intact] {%
    \label{fig:experiments:realistic:amputeeintact}%
    \tikzsetfigurename{adapt_realistic_AI}
    \begin{tikzpicture}
      \begin{axis}[yticklabel=\empty,legend to name=fig:experiments:realistic:legend]
        \addplot+[notransfer] table[] {acc_no_AI_realistic.txt};
        \addlegendentry{\acs{NOTRANSFER}}
        \addplot+[priorlin] table[] {acc_prior_AI_realistic.txt};
        \addlegendentry{\acs{PRIORLIN}}
        \addplot+[multikt] table[] {acc_multikt_AI_realistic.txt};
        \addlegendentry{\acs{MULTIKT}}
      \end{axis}
    \end{tikzpicture}
  }%
  \caption{Standard classification accuracy for \ac{MULTIKT}, \ac{NOTRANSFER} and \ac{PRIORLIN} in the realistic setting.}%
  \label{fig:experiments:realistic}%
\end{figure*}
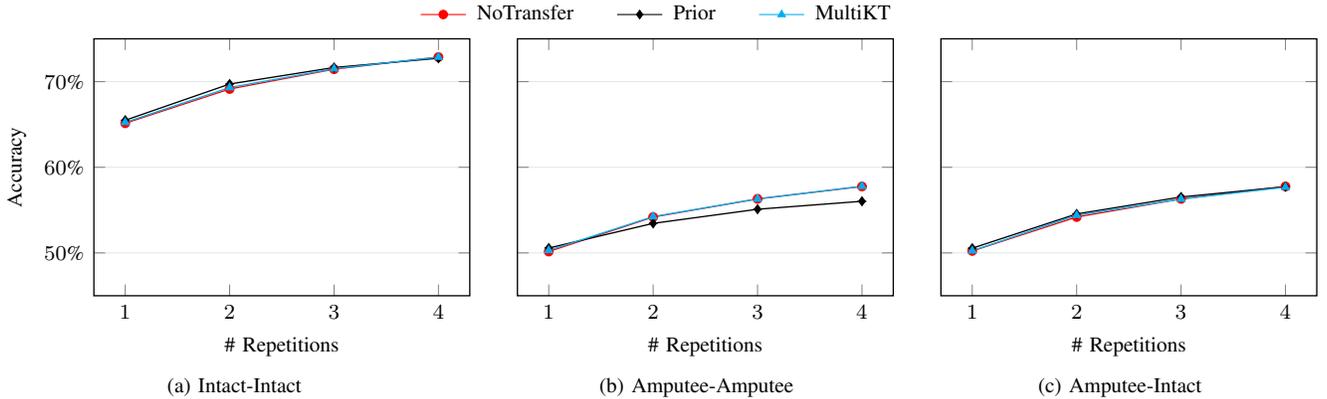

\autoref{fig:experiments:realistic} shows the standard classification accuracy for the realistic setting described in \autoref{sec:setup:settings:realistic} averaged over the target subjects and all possible combinations of a given number of training repetitions. 
Also in this setting the hyperparameters were tuned appropriately and the differences among the methods are again negligible.
In addition, we observe significantly lower accuracy among amputees compared to intact subjects, confirming the deterioration of the myoelectric signals due to amputation and subsequent lack of muscular use.

\subsection{Discussion}\label{sec:experiments:discussion}


The results clearly show that the improvements usually attributed to prior knowledge can instead be explained by suboptimal hyperparameter optimization. 
With properly tuned hyperparameters, the \ac{NOTRANSFER} baseline that completely ignores source information performs as well as the more complicated domain adaptation methods.

There are multiple explanations for the observed differences in performance in the original setting. 
First, the hyperparameters were chosen based on the performance of an \ac{SVM} when transferring from the source subjects to the target subjects.
This gives a disadvantage to \ac{NOTRANSFER}, which does not exploit prior knowledge to train the classifier. 
Furthermore, this parameter setting is also problematic since all methods are based on \ac{LSSVM}, which uses a different loss function than \ac{SVM}. 
As can be seen in the objective function in \autoref{eq:ls-svm}, the regularization parameter $C$ is multiplied with the absolute magnitude of all training losses, so an optimal setting for \ac{SVM} does not necessarily work well for \ac{LSSVM}.

A further problem is that the value of $C$ was determined using the total set of training samples.
The same value was subsequently used when training on much smaller subsets, leading to a different tradeoff between minimizing training errors and regularizing the solution.
This affects all methods except \ac{MKAL}, for which the specific implementation multiplied the given value of $C$ with the number of training samples.
\ac{MKAL} therefore effectively used a much larger value of $C$ (i.e, much less regularization), explaining why it performed superior to the other methods.

A similar, though slightly more complicated, argument holds for \ac{MULTIKT}. 
Recall the formulation of biased regularization in \autoref{eq:OptimMultiAdaMulti}; the linear combination of source hypotheses allows to reduce the effect of the regularization term by moving the bias in the direction of the optimal solution. 
In other words, for the same value of $C$ this allows to concentrate more on minimizing the training errors on the target data.

%% file: Conclusions.tex
\section{Conclusions} \label{sec:conclusions}

In this paper, we have tested two popular domain adaptation algorithms that were proposed to reduce the training time needed to control a prosthesis.
We found that the improvements in earlier studies can in fact be attributed to suboptimal hyperparameter optimization, which penalized in particular the \ac{NOTRANSFER} reference method.
When the hyperparameters are appropriately tuned on the training data of the target subject, the previously reported differences vanish.

This result also holds when correcting for other technical and conceptual mistakes in the original experimental framework.
The accuracy of the classification methods in our updated setting was evaluated with respect to the number of repetitions of each movement, which represents the real effort for the user during the training phase, and for all subjects in the \ac{NINAPRO} database.
Also in this case, we do not observe any differences between the \ac{HTL} algorithms and the \ac{NOTRANSFER} baseline.

Intuitively, it should be possible to improve performance on a specific task by using prior information from related tasks. 
Our findings show, however, that in the context of prosthetic control \ac{MULTIKT} and \ac{MKAL}, which transfer just source hypotheses rather than source data, do not lead to improved performance. 
In future work, we will therefore continue to investigate how to successfully leverage over prior information to reduce the training effort for an amputee. Among the directions we consider are unsupervised domain adaptation via distribution alignment~\cite{pan11,fernando13} and subject invariant data representations using deep learning methods.

%% file: icorr17.bbl
\begin{thebibliography}{24}
\providecommand{\natexlab}[1]{#1}
\providecommand{\url}[1]{#1}
\csname url@samestyle\endcsname
\providecommand{\newblock}{\relax}
\providecommand{\bibinfo}[2]{#2}
\providecommand{\BIBentrySTDinterwordspacing}{\spaceskip=0pt\relax}
\providecommand{\BIBentryALTinterwordstretchfactor}{4}
\providecommand{\BIBentryALTinterwordspacing}{\spaceskip=\fontdimen2\font plus
\BIBentryALTinterwordstretchfactor\fontdimen3\font minus
  \fontdimen4\font\relax}
\providecommand{\BIBforeignlanguage}[2]{{%
\expandafter\ifx\csname l@#1\endcsname\relax
\typeout{** WARNING: IEEEtranN.bst: No hyphenation pattern has been}%
\typeout{** loaded for the language `#1'. Using the pattern for}%
\typeout{** the default language instead.}%
\else
\language=\csname l@#1\endcsname
\fi
#2}}
\providecommand{\BIBdecl}{\relax}
\BIBdecl

\bibitem[Atkins et~al.(1996)Atkins, Heard, and Donovan]{atkins96}
D.~J. Atkins, D.~C.~Y. Heard, and W.~H. Donovan, ``Epidemiologic overview of
  individuals with upper-limb loss and their reported research priorities,''
  \emph{Journal Of Prosthetics And Orthotics}, vol.~8, no.~1, pp. 2--11, 1996.

\bibitem[Castellini et~al.(2009)Castellini, Fiorilla, and
  Sandini]{castellini09}
C.~Castellini, A.~E. Fiorilla, and G.~Sandini, ``Multi-subject / daily-life
  activity {EMG}-based control of mechanical hands,'' \emph{Journal of
  Neuroengineering and Rehabilitation}, vol.~6, no.~41, 2009.

\bibitem[Orabona et~al.(2009)Orabona, Castellini, Caputo, Fiorilla, and
  Sandini]{orabona09}
F.~Orabona, C.~Castellini, B.~Caputo, A.~E. Fiorilla, and G.~Sandini, ``Model
  adaptation with least-square svm for adaptive hand prosthetics,'' in
  \emph{IEEE International conference on Robotics and Automation}, 2009.

\bibitem[Tommasi et~al.(2013)Tommasi, Orabona, Castellini, and
  Caputo]{tommasi11}
T.~Tommasi, F.~Orabona, C.~Castellini, and B.~Caputo, ``Improving control of
  dexterous hand prostheses using adaptive learning,'' \emph{{IEEE}
  Transactions on Robotics}, vol.~29, no.~1, pp. 207--219, 2013.

\bibitem[Patricia et~al.(2014)Patricia, Tommasi, and Caputo]{patricia14}
N.~Patricia, T.~Tommasi, and B.~Caputo, ``Multi-source adaptive learning for
  fast control of prosthetics hand,'' in \emph{2014 {IEEE} Conference on
  Computer Vision and Pattern Recognition {CVPR}}, 6 2014, pp. 2769--2774.

\bibitem[Farina et~al.(2002)Farina, Cescon, and Merletti]{farina02}
D.~Farina, C.~Cescon, and R.~Merletti, ``Influence of anatomical, physical, and
  detection-system parameters on surface {EMG},'' \emph{Biological
  Cybernetics}, vol.~86, no.~6, pp. 445--456, 2002.

\bibitem[Graupe and Cline(1975)]{graupe75}
D.~Graupe and W.~K. Cline, ``Functional separation of {EMG} signals via {ARMA}
  identification methods for prosthesis control purposes,'' \emph{IEEE
  Transactions on Systems, Man and Cybernetics}, no.~2, pp. 252--259, 1975.

\bibitem[Matsubara et~al.(2011)Matsubara, Hyon, and Morimoto]{matsubara11}
T.~Matsubara, S.~Hyon, and J.~Morimoto, ``Learning and adaptation of a
  stylistic myoelectric interface: {EMG}-based robotic control with individual
  user differences,'' in \emph{{ROBIO}}.\hskip 1em plus 0.5em minus 0.4em\relax
  IEEE, 2011, pp. 390--395.

\bibitem[Sensinger et~al.(2009)Sensinger, Lock, and Kuiken]{sensingero09}
J.~W. Sensinger, B.~A. Lock, and T.~A. Kuiken, ``Adaptive pattern recognition
  of myoelectric signals: Exploration of conceptual framework and practical
  algorithms,'' \emph{IEEE Transactions on Neural systems and rehabilitation
  engineering}, vol.~17, no.~3, 2009.

\bibitem[Chattopadhyay et~al.(2011)Chattopadhyay, Krishnan, and
  Panchanathan]{chattopadhyay11}
R.~Chattopadhyay, N.~Krishnan, and S.~Panchanathan, \emph{Topology preserving
  domain adaptation for addressing subject based variability in SEMG signal},
  2011, vol. SS-11-04, pp. 4--9.

\bibitem[Khushaba(2014)]{khushaba14}
R.~N. Khushaba, ``Correlation analysis of electromyogram signals for multiuser
  myoelectric interfaces,'' \emph{IEEE Transactions on Neural Systems and
  Rehabilitation Engineering}, vol.~22, no.~4, pp. 745--755, 2014.

\bibitem[Liu et~al.(2016{\natexlab{a}})Liu, Sheng, Zhang, Jiang, and
  Zhu]{liu16towards}
J.~Liu, X.~Sheng, D.~Zhang, N.~Jiang, and X.~Zhu, ``Towards zero retraining for
  myoelectric control based on common model component analysis,'' \emph{IEEE
  Transactions on Neural Systems and Rehabilitation Engineering}, vol.~24,
  no.~4, pp. 444--454, 2016.

\bibitem[Liu et~al.(2016{\natexlab{b}})Liu, Sheng, Zhang, He, and Zhu]{liu16}
J.~Liu, X.~Sheng, D.~Zhang, J.~He, and X.~Zhu, ``Reduced daily recalibration of
  myoelectric prosthesis classifiers based on domain adaptation,'' \emph{IEEE
  journal of biomedical and health informatics}, vol.~20, no.~1, pp. 166--176,
  2016.

\bibitem[Atzori et~al.(2014)Atzori, Gijsberts, Castellini, Caputo, Hager,
  Elsig, Giatsidis, Bassetto, and M{\"u}ller]{atzori14}
M.~Atzori, A.~Gijsberts, C.~Castellini, B.~Caputo, A.-G.~M. Hager, S.~Elsig,
  G.~Giatsidis, F.~Bassetto, and H.~M{\"u}ller, ``Electromyography data for
  non-invasive naturally-controlled robotic hand prostheses,'' \emph{Scientific
  data}, vol.~1, 2014.

\bibitem[Suykens et~al.(2002)Suykens, Van~Gestel, De~Brabanter, De~Moor, and
  Vandewalle]{suykens02}
J.~Suykens, T.~Van~Gestel, J.~De~Brabanter, B.~De~Moor, and J.~Vandewalle,
  \emph{Least Squares Support Vector Machines}.\hskip 1em plus 0.5em minus
  0.4em\relax World Scientific, 2002.

\bibitem[Tommasi et~al.(2014)Tommasi, Orabona, and Caputo]{tommasi14}
T.~Tommasi, F.~Orabona, and B.~Caputo, ``Learning categories from few examples
  with multi model knowledge transfer,'' \emph{{IEEE} Trans. Pattern Anal.
  Mach. Intell.}, vol.~36, no.~5, pp. 928--941, 2014.

\bibitem[Orabona et~al.(2010)Orabona, Luo, and Caputo]{orabona10}
F.~Orabona, J.~Luo, and B.~Caputo, ``Online-batch strongly convex multi kernel
  learning,'' in \emph{{IEEE} Conference on Computer Vision and Pattern
  Recognition}, June 2010, pp. 787--794.

\bibitem[Orabona et~al.(2012)Orabona, Luo, and Caputo]{orabona12}
------, ``Multi kernel learning with online-batch optimization,'' \emph{Journal
  of Machine Learning Research}, vol.~13, pp. 227--253, 2012.

\bibitem[Englehart and Hudgins(2003)]{englehart03}
K.~Englehart and B.~Hudgins, ``A robust, real-time control scheme for
  multifunction myoelectric control,'' \emph{IEEE Transactions on Biomedical
  Engineering}, vol.~50, no.~7, pp. 848--854, 2003.

\bibitem[Kuzborskij et~al.(2012)Kuzborskij, Gijsberts, and
  Caputo]{kuzborskij12}
I.~Kuzborskij, A.~Gijsberts, and B.~Caputo, ``On the challenge of classifying
  52 hand movements from surface electromyography,'' in \emph{Annual
  International Conference of the IEEE Engineering in Medicine and Biology
  Society}, 2012, pp. 4931--4937.

\bibitem[Kuzborskij and Orabona(2016)]{kuzborskij16}
I.~Kuzborskij and F.~Orabona, ``Fast rates by transferring from auxiliary
  hypotheses,'' \emph{Machine Learning}, pp. 1--25, 2016.

\bibitem[Gijsberts et~al.(2014)Gijsberts, Atzori, Castellini, M{\"u}ller, and
  Caputo]{gijsberts14}
A.~Gijsberts, M.~Atzori, C.~Castellini, H.~M{\"u}ller, and B.~Caputo,
  ``Movement error rate for evaluation of machine learning methods for
  s{EMG}-based hand movement classification,'' \emph{IEEE Transactions on
  Neural Systems and Rehabilitation Engineering}, vol.~22, no.~4, pp. 735--744,
  2014.

\bibitem[Pan et~al.(2011)Pan, Tsang, Kwok, and Yang]{pan11}
S.~J. Pan, I.~W. Tsang, J.~T. Kwok, and Q.~Yang, ``Domain adaptation via
  transfer component analysis,'' \emph{IEEE Transactions on Neural Networks},
  vol.~22, no.~2, pp. 199--210, Feb 2011.

\bibitem[Fernando et~al.(2013)Fernando, Habrard, Sebban, and
  Tuytelaars]{fernando13}
B.~Fernando, A.~Habrard, M.~Sebban, and T.~Tuytelaars, ``Unsupervised visual
  domain adaptation using subspace alignment,'' in \emph{2013 IEEE
  International Conference on Computer Vision}, Dec 2013, pp. 2960--2967.

\end{thebibliography}
